\documentclass[a4paper, 10pt, conference]{ieeeconf}      

\IEEEoverridecommandlockouts                              

\usepackage{url}
\usepackage{comment}
\usepackage{amsmath,amssymb,amsfonts, mathtools}
\usepackage{graphicx}
\usepackage{bm}
\usepackage{epstopdf}
\usepackage{placeins}
\usepackage{bm}
\usepackage{cite}
\let\labelindent\relax
\usepackage[inline]{enumitem}
\usepackage{multicol}
\usepackage{multirow}
\usepackage{tikz}
\usepackage{caption}
\usepackage{subcaption}
\usepackage[hidelinks]{hyperref}
\usepackage{siunitx}
\usepackage{gensymb}
\usepackage{supertabular}
\usepackage{booktabs}
\newcommand\reza{\textcolor{red}}

\newcommand{\vect}[1]{\boldsymbol{#1}}
\newcommand\numberthis{\addtocounter{equation}{1}\tag{\theequation}}

\definecolor{Green_table}{RGB}{60,179,113}
\definecolor{Red_table}{RGB}{255,0,0}
\definecolor{Orange_table}{RGB}{255,140,0}

\title{\LARGE \bf An Adaptive Framework for Manipulator Skill Reproduction in Dynamic Environments}

\author{Ryan Donald,$^1$ Brendan Hertel,$^1$ Stephen Misenti,$^{1,2}$ Yan Gu$^2$ and Reza Azadeh$^1$
	\thanks{$^1$ Persistent Autonomy and Robot Learning (PeARL) Lab, University of Massachusetts Lowell, MA, 01854. Email: \texttt{\{ryan\_donald, brendan\_hertel, stephen\_misenti\}@student.uml.edu, reza@cs.uml.edu} \newline
    $^2$ Terrain Robotics Advanced Control and Experimentation (TRACE) Lab, Purdue University, IN, 47907. Email: \texttt{yangu@purdue.edu}}
}

\begin{document}

\maketitle

\begin{abstract}
    Robot skill learning and execution in uncertain and dynamic environments is a challenging task. This paper proposes an adaptive framework that combines Learning from Demonstration (LfD), environment state prediction, and high-level decision making. Proactive adaptation prevents the need for reactive adaptation, which lags behind changes in the environment rather than anticipating them. We propose a novel LfD representation, Elastic-Laplacian Trajectory Editing (ELTE), which continuously adapts the trajectory shape to predictions of future states. Then, a high-level reactive system using an Unscented Kalman Filter (UKF) and Hidden Markov Model (HMM) prevents unsafe execution in the current state of the dynamic environment based on a discrete set of decisions. We first validate our LfD representation in simulation, then experimentally assess the entire framework using a legged mobile manipulator in 36 real-world scenarios. We show the effectiveness of the proposed framework under different dynamic changes in the environment. Our results show that the proposed framework produces robust and stable adaptive behaviors.
\end{abstract}

\section{Introduction}
\label{sec:intro}

As robots become intertwined with human environments, they must robustly negotiate the difficulties of these environments. It may be easy to navigate and manipulate in a structured warehouse, but with clutter and debris the complexity of such a task increases. One domain which has been under-explored is manipulation in dynamic environments that are characterized by time-varying changes in the robot's surroundings. Such perturbations in the environment could be caused by moving targets and obstacles. Another important factor that can result in unstructured and dynamic environments is the movement of the ground in the inertial frame. This type of movement has been investigated in the control of legged locomotion~\cite{9108552,9847283,iqbal2021extended,gao2023time}. While achieving reliable locomotion in such environments remains a challenging problem, mobile manipulation in such environments has shown to be complex and poses many issues~\cite{schmitt2019planning}. First, the base must be stabilized, and a manipulator should be able to smoothly execute the task despite perturbations in the base. Additionally, a reactive system should be operating to provide the robot with the ability to react and safely avoid obstacles if the perturbations increase dramatically. 

\begin{figure}[ht]
    \centering
    \includegraphics[width=0.6\linewidth]{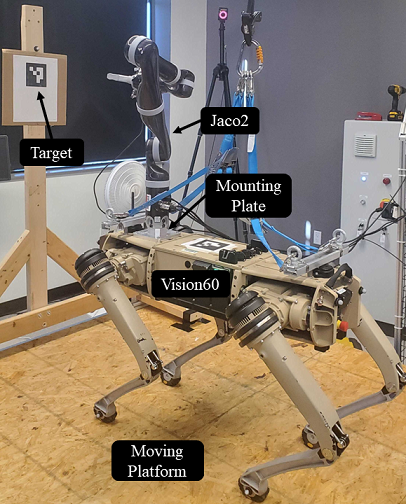}
    \caption{\small Experimental setup with Kinova Jaco2 mounted on a Ghost Robotics Vision 60.}
    \label{fig:setup}
\end{figure}

To achieve safe and robust robot manipulation in dynamic environments, we propose an adaptive skill learning framework consisting of three main modules. First, a novel Learning from Demonstration (LfD) representation is proposed that can quickly adapt online to perturbations in the environment. Our LfD representation, Elastic-Laplacian Trajectory Editing (ELTE), combines ideas from Laplacian Trajectory Editing (LTE)~\cite{Nierhoff2016LTE} and elastic maps~\cite{hertel2022ElMap}. This representation is able to smoothly deform trajectories online while maintaining the shapes of demonstrated skill. The next part of the framework consists of a state estimation module, using the Unscented Kalman Filter (UKF)~\cite{wan2000unscented}, that can provide prediction of the environment state at a future time resulting in a proactive adaptation. This module allows the framework to generate smoother reproductions, as changes in the environment are predicted and incorporated into execution before happening. The final part of the framework is a high-level decision making system utilizing a Hidden Markov Model (HMM)~\cite{rabiner1989hmm} that allows the robot to react to changes in the state of the environment to avoid collision and adapt the robot movement when the environment is unstable or otherwise unsuitable for execution. To validate our framework, we use a legged mobile manipulator system that combines a Kinova Jaco2 arm and a Ghost Robotics Vision 60 legged base (shown in Fig~\ref{fig:setup}). Using this system, we conduct experiments including inspecting a moving target when the base is static and inspecting a target marker when the base is self-stabilizing on a dynamic floor. 

\section{Related Work}
\label{sec:RW}

There are a variety of approaches for online adaptation for manipulators in robotics. In learning from demonstration, previous approaches have shown their robustness against perturbations~\cite{Argall2009survey}. An LfD representation which is robust against perturbations will still continue execution even if the robot or goal is perturbed during execution. This is usually achievable because the LfD representation is modeled as a dynamical system~\cite{pastorDMP2009, Khansari-Zadeh2011LASA}, and the system dynamics change with the environment. In some cases, the goal is perturbed, and the dynamical system incorporates the goal and adapts accordingly, such as Dynamic Movement Primitives (DMPs)~\cite{pastorDMP2009}. In other cases, a stable dynamic system like Stable Estimator of Dynamical Systems (SEDS)~\cite{Khansari-Zadeh2011LASA} is used, allowing the robot to return to the path after a perturbation. These LfD systems provide only trajectory adaptation based on current environment information, and have no ability to proactively adjust the trajectory, anticipate a future change in the endpoint, or react to the environment to prevent unsafe execution. Some works have included a reactive system on top of DMPs such as \cite{ahmadzadeh2014reactiveValve}, which reacts to unsafe execution using a fuzzy decision maker. However, this work still does not present a proactive solution to changes in the environment.

Other approaches use a variety of control policies for reactive behavior. A common method of control is visual servoing~\cite{kragic2002survey}, where image or video feedback is used to control robot execution. Often, images are used to estimate the state of the robot and its environment which informs a separate robot controller that handles execution in the environment. Visual servoing can be a simple yet powerful tool, but other techniques can provide more information. For example, \cite{abidi1991autonomous} uses multiple sensors such as vision, proximity, and force/torque to semi-autonomously execute an inspection task in a variable environment. While this system reacts to different environments, it does not react to disturbances in the environment during execution.


Additionally, work has been done to incorporate the non-inertial motion of a platform into robot control systems. A simple form of this is shown in \cite{from2009}, where the motion of the non-inertial platform is included in the formulation of the control system, allowing for efficient control of a robot when compared to standard methods, as long as the non-inertial platform's motion is known. This was then expanded upon in \cite{from2011}, where a predictive measure is used to predict the motion of the platform, in this case a ship. They propose an auto-regressive predictor, as well as a superposition of sine waves as two methods to provide the prediction of the ship's motion. This is shown to improve the control of the system, as long as the motion is predictable.

Few methods incorporate proactive movements, or predictive reactive movements. Proactive human-robot collaboration has been proposed as a future manufacturing paradigm where a robot is proactively planning movements~\cite{li2021towards}. However, for predictive control of manipulators, especially in dynamic environments, few works exist. Woolfrey et al.~\cite{woolfrey2021predictive} create a model for disturbances in the environment, and adjust execution based on the model. This is limited to environments with motion that can easily be modeled, such as periodic motion. Our framework uses a more general model to predict future movements, and does not require disturbances to be periodic. Additionally, we can adapt online to changes in the environment and react to the environment to prevent unsafe execution.

\begin{figure}[ht]
    \centering
    \includegraphics[width=0.98\columnwidth]{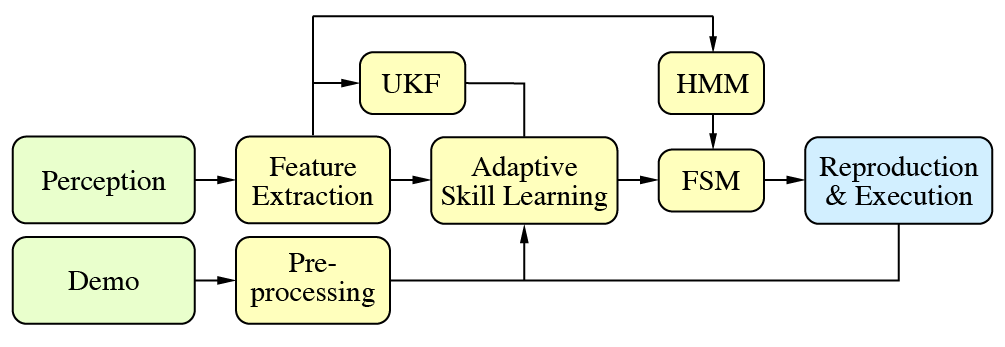}
    \caption{\small An overview of all components of the proposed adaptive skill learning and execution framework. }
    \label{fig:framework}
\end{figure}

\section{Methodology}
\label{sec:method}

The proposed adaptive skill learning framework must be able to deal with the unstable and dynamic changes in the environment. In other words, such a framework must be able to perform not only based on its observations, but it must be able to predict the environment based on previous and current information. 

As depicted in Fig~\ref{fig:framework}, the proposed framework includes several modules. 
First, cameras and motor encoders are used to perceive the current state of the dynamic environment. From this perception, we extract important features from the environment. In the case of manipulation on a dynamic environment, these features are the current robot and goal states. These features are then used as input to several other modules. To predict the state of the environment in future timesteps, we include a state estimation module, namely, the Unscented Kalman Filter (UKF)~\cite{wan2000unscented}. This prediction of the state is used to generate/update the skill execution. We develop an adaptive skill learning representation to encode and reproduce trajectories that also allows for online adaptation of the movement according to the changing states and predictions. To react to the perturbations in the environment in high-level, we design a Hidden Markov Model (HMM)~\cite{rabiner1989hmm}. This HMM uses the changing state of the environment to determine if it is safe to execute the encoded skill in the current state. If execution is considered safe, execution proceeds normally. Otherwise, the robot reacts according to the environment state by halting or retracting from the goal. We develop a  Finite State Machine (FSM) to handle the decision making of the HMM.  


\subsection{Adaptive Skill Learning}
We propose a novel skill learning from demonstration method, Elastic-Laplacian Trajectory Editing (ELTE),\footnote{Available at: \url{https://github.com/brenhertel/ELTE}} that combines ideas from elastic maps~\cite{hertel2022ElMap} and Laplacian Trajectory Editing (LTE)~\cite{Nierhoff2016LTE}. Elastic maps are created by finding a set of nodes $\vect{y}$ which represent the data, and reduce the stretching and bending energies of the ``spring'' connections between nodes. Overall, an elastic map is found by minimizing three energies: 
\begin{enumerate*}[label=(\roman*)]
  \item the approximation energy $U_Y$ which penalizes a bad fit to the data,
  \item the stretching energy $U_E$ which penalizes high distance between adjacent nodes, and
  \item the bending energy $U_R$ which penalizes the curvature of nodes.
\end{enumerate*} 
The constructed elastic map is then used for movement reproduction, as it has desirable features such as smoothness, incorporation of initial, final, or via-point constraints, and can generalize to one or more demonstrations. In this work, we modify the approximation energy, $U_Y$, to change how trajectories adapt, as well as the optimization method of the map, resulting in an adaptive skill learning representation. With the original elastic map optimization, reproductions are rewarded for following the given demonstration. However, we wish to deform trajectories such that they maintain the shape of the given demonstration and have no incentive for converging to the given demonstration, as in trajectory editing methods~\cite{Nierhoff2016LTE}. Additionally, it is necessary that we optimize the map during execution, and previously executed portions of the map must not be changed. Therefore, we modify the optimization of the map to allow for online adaptation.


We define a demonstration $\vect{\zeta}$ as a vector of points $[ \zeta_1, \zeta_2, ..., \zeta_T]^\top$ with an individual $d$-dimensional point $\zeta_i$. A reproduction, $\vect{y} = [ y_1, y_2, ..., y_T]^\top$, is found by minimizing the energy objectives listed above. We use a convex formulation for efficient optimization with flexible constraints~\cite{hertel2023confidence}. These convex objectives are formulated as 
\begin{align}
    U_Y &= || \vect{L}\vect{y} -\vect{L}\vect{\zeta} ||_2^2 \label{eq:U_Y_new}\\
    U_E &= w_E  || \vect{E}\vect{y} ||_2^2  \label{eq:U_E_new}\\
    U_R &= w_R  || \vect{R}\vect{y} ||_2^2, \label{eq:U_R_new}
\end{align}
where $w_E, w_R$ are weight parameters, $||\cdot||_n$ is the $L^n$-norm, $\vect{L}$ is the graph Laplacian~\cite{Nierhoff2016LTE}, and the matrices $\vect{E}$ and $\vect{R}$ are the first and second order finite difference matrices, respectively~\cite{hertel2023confidence}. The convex optimization involving these constraints is formulated as 
\begin{align*}
    &\underset{\vect{y}}{\text{minimize }}  f_0(\vect{y}) = U_Y + U_E + U_R \numberthis\label{eq:opt} \\
    &\text{subject to} \ f_i(\vect{y}) = ||p - y_j||_1 - r \leq 0
\end{align*}
where $f_i(\vect{y})$ is the $i$th constraint, constraining some point of the reproduction $y_j$ within a radius $r$ of a point $p$. The inequality constraint here is used as the perception system may have low confidence in the goal position until later in the trajectory. Therefore, we can begin modifying the trajectory early with large radii and refine the constraint as the trajectory executes. Additionally, using this formulation we can provide other constraints such as obstacles and via-points if necessary (see \cite{hertel2023confidence} for details).

According to the context of the environment (i.e., output of the perception and feature extraction systems that detects the target object), the encoded movement must be generalized to new situations. In this paper, we explore moving targets, meaning that the endpoint must be modified. Therefore, during execution, we modify the trajectory online. Given that we have already executed some fraction of the trajectory for timesteps $0:t$, we must modify the rest of the trajectory for timesteps $t+1:T$. Therefore, we modify \eqref{eq:opt} for online adaptation as
\begin{align*}
    &\underset{\vect{y}_{t+1:T}}{\text{minimize }}  f_0(\vect{y}) = U_Y + U_E + U_R \numberthis\label{eq:online_opt} \\
    &\text{subject to} \ f_1(\vect{y}) = ||p - y_T||_1 - r \leq 0
\end{align*}
where here we specify $p$ as the current prediction of the target final location and $r$ as some region around that target. Note that while $\vect{y}_{0:t}$ is not included in the solution to the problem, they are still included in the optimization as they affect the energies associated with the map. The updated problem formulation remains convex and can be solved efficiently, allowing for smooth online adaptation which maintains the shape of given demonstrations. Additionally, to increase the efficiency of optimizing the elastic map, we do not use the clustering process and Expectation-Maximization (EM) algorithm as shown in \cite{hertel2022ElMap}. Instead, we skip the clustering step by connecting each node to a corresponding data point from the demonstration, then solving \eqref{eq:online_opt}. In fact, EM would be unable to solve the online adaptation problem, as it optimizes the complete map instead of a portion.

\subsection{Environment Prediction}

For short-term prediction, we utilize the Unscented Kalman Filter (UKF)~\cite{wan2000unscented}. This is done through a short term linear approximation of the input, similar to the Extended Kalman Filter (EKF)~\cite{gustafsson2011ekfukf}, with the exception of using sigma points to provide a more accurate estimation. In our framework, features of the environment's state are input to the UKF, and a prediction is generated for a future time-step. As a short-term prediction, this method allows for accurate prediction of the environment's movement, but also allows for prediction at variable time-steps in the future with different levels of certainty. The longer-term prediction provides the robot with a general goal at the beginning of execution which becomes more refined over the execution duration.

The UKF can be defined with the steps below. First, we define the state space, $\vect{x}$, as a vector of $m$ features, in our case this is the vector $\vect{x} = [x_{\hat{i}}, x_{\hat{j}}, x_{\hat{k}}, \dot{x}_{\hat{i}}, \dot{x}_{\hat{j}}, \dot{x}_{\hat{k}}]$, where $x$ is the position of the target and $\dot{x}$ is the velocity of the target. Similarly we have the observation, $\vect{z}$, which is the current position of the marker in the coordinate frame of the robot. Alongside this, the process and observation models at time step $k$ are defined as $\vect{x}_k = F(\vect{x}_{k-1}) + \vect{v}_k$ and $\vect{z}_k = H(\vect{x}_k) + \vect{n}_k$, respectively. These equations introduce noise to the process and observation models, where $\vect{v}_k$ is the process noise and $\vect{n}_k$ is the observation noise. The process model $F$ is defined as a constant-acceleration kinematics system along each axis, with the observation model $H$ as the measured position. We then define the sigma-points matrix $\vect{\chi}$ as a $2m + 1$ matrix, where $m$ is the number of dimensions in the state vector, as a parameterized set of sigma points, where sigma point $\chi_i$ has weight $W_i$. For full details, see \cite{wan2000unscented}. 

The state estimations from the UKF can be used to predict future states at each timestep. For target tracking, the final timestep $T$ is predicted, where $\vect{x}_T = [y_T, \dot{y}_T]$. This prediction is constantly updaed with each new observation, thus updating the trajectory generated by our adaptive skill learning module formulated by \eqref{eq:online_opt}.


\begin{figure}[ht]
    \centering
    \includegraphics[width=0.98\linewidth]{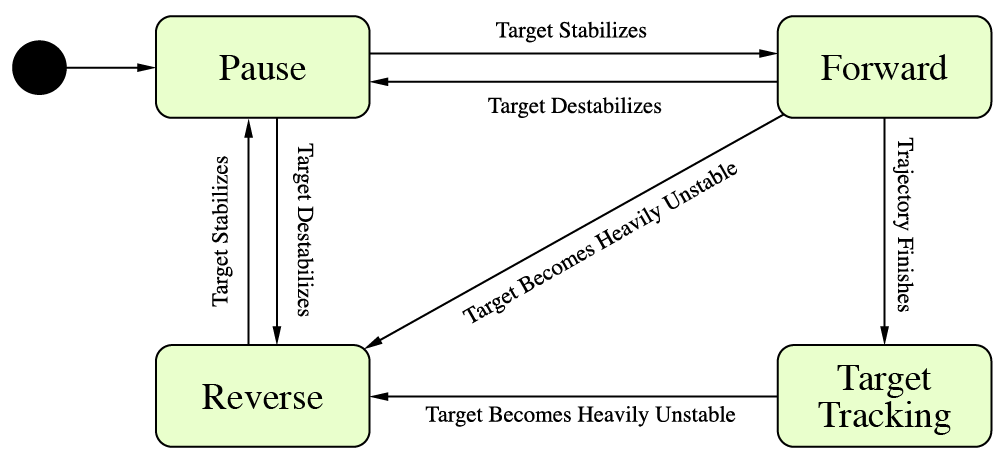}
    \caption{\small Finite State Machine (FSM) that controls high-level state transition based on changes in the environment.}
    \label{fig:fsm}
\end{figure}

\subsection{Reactive System and High-Level Decision Making}

To allow higher-level decision making, we utilize a Hidden Markov Model (HMM)~\cite{rabiner1989hmm}. We define this as a model with the following discrete set of hidden states: $ S = [\textit{Forward}, \textit{Pause}, \textit{Reverse}]$ and continuous set of observable states $V = [\dot{x}]$. These states are chosen as to appropriately react to the velocity of the target. If the target is moving too fast, it is likely unstable, and execution should either \textit{Pause} or \textit{Reverse} for safety. Once the velocity slows, trajectory execution can continue as normal, returning to the \textit{Forward} state.

Given the change in state from the HMM, we design a Finite State Machine (FSM) to govern execution, shown in Fig.~\ref{fig:fsm}. The FSM receives the trajectory from the skill reproduction module and the current state of execution from the HMM. The FSM internally records where in the duration of execution the current time-step is, and can use these time-steps to either move forward, pause, or reverse execution. If perturbations in the environment are low, execution continues as normal. However, if perturbations are too high and considered unsafe according to the trained HMM, execution can either be paused or reversed. If execution is reversed, the trajectory generation overwrites previous execution to allow for safer and new adaptive motion. Once execution finishes, the FSM moves into a ``target tracking'' state where the manipulator continues to follow the target. However, the robot can still exit out of this state and begin retracting from the target if it becomes unstable.

\section{Experiments}
\label{sec:exps}

\subsection{Validation of the Adaptive Skill Learning in Simulation}

\begin{figure}[t]
    \centering
    \includegraphics[width=0.32\linewidth]{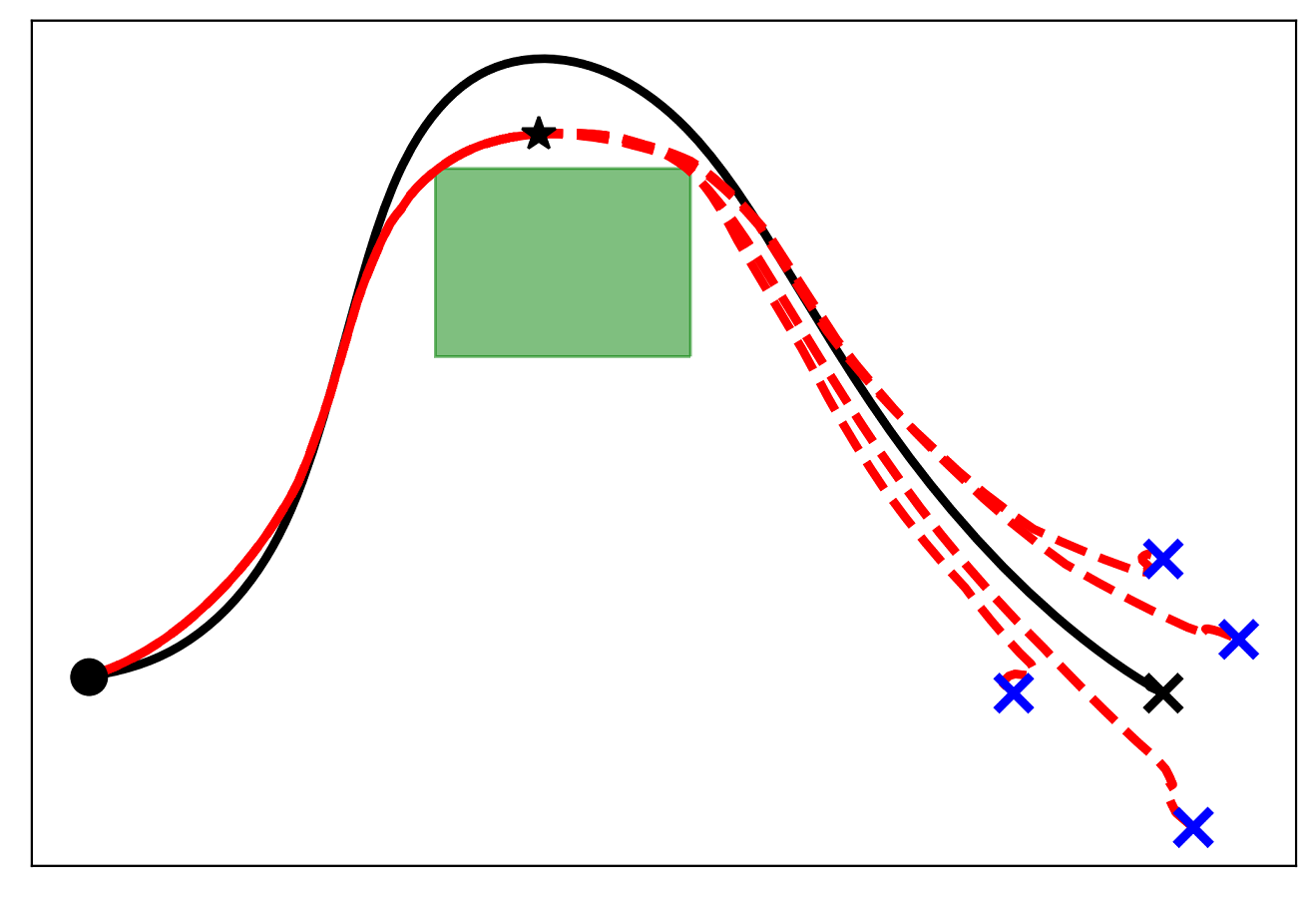}
    \includegraphics[width=0.32\linewidth]{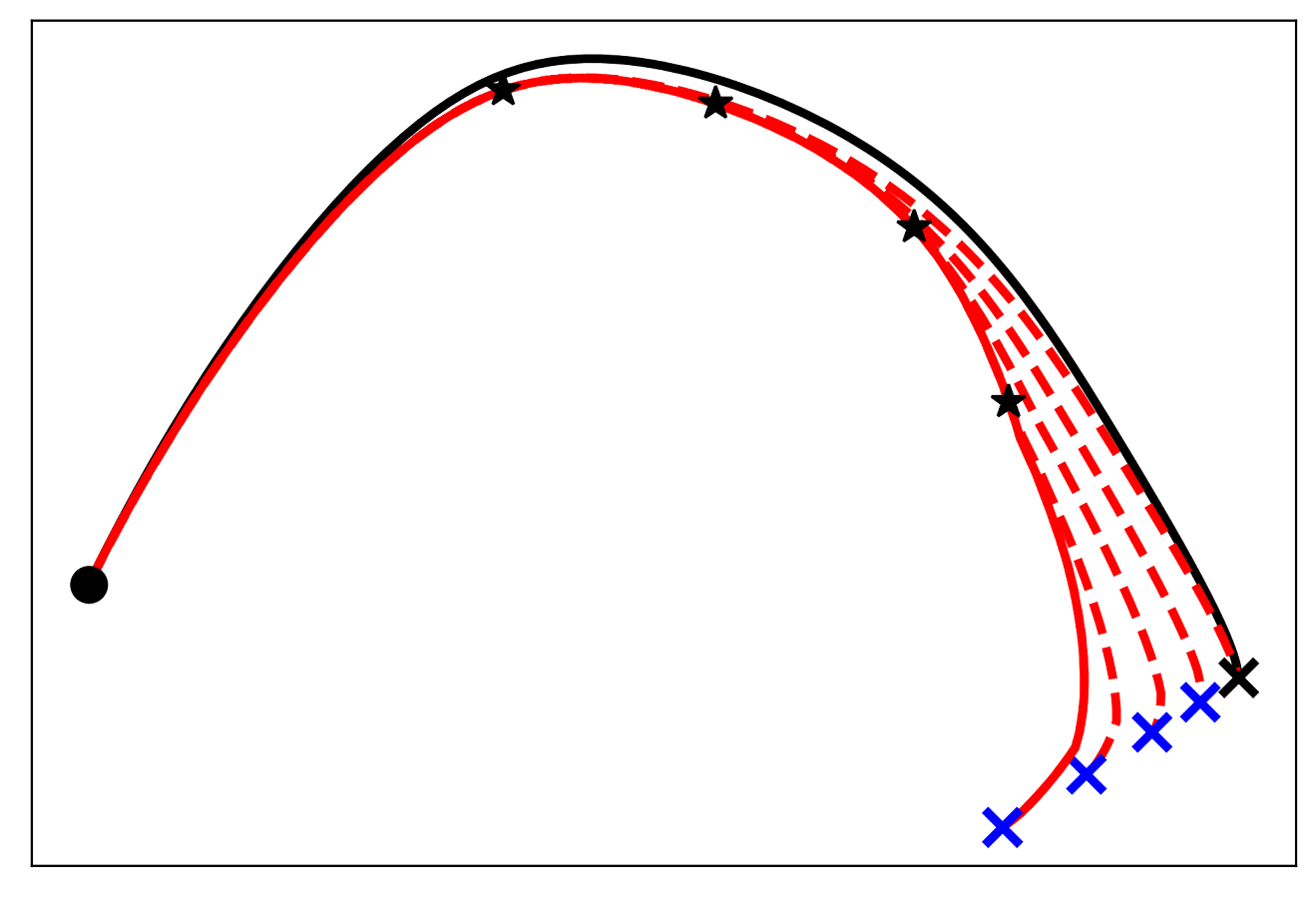}
    \includegraphics[width=0.32\linewidth]{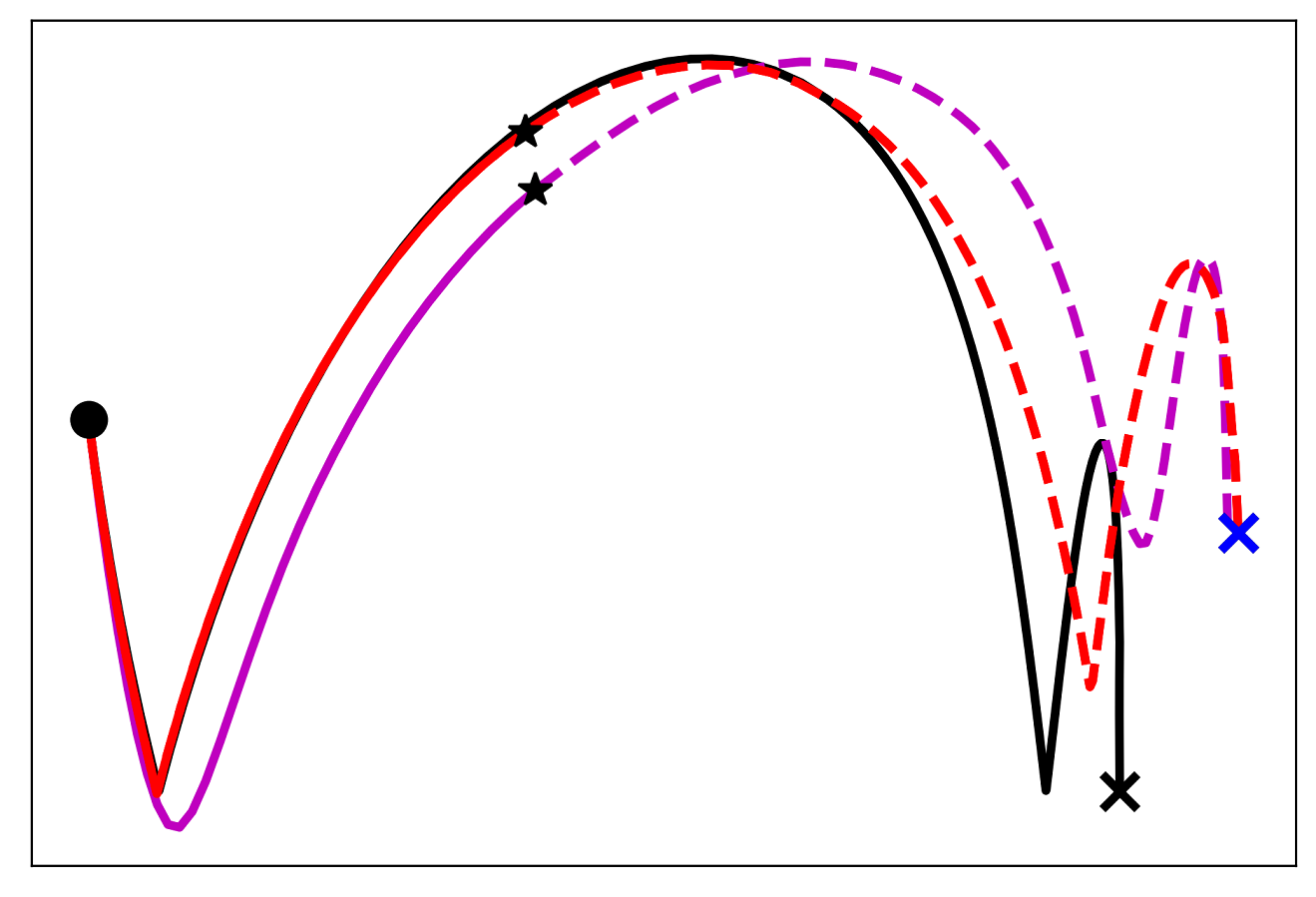}
    \includegraphics[width=0.98\linewidth]{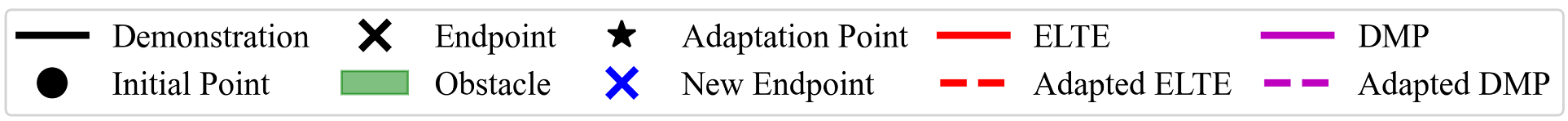}
    \caption{\small (Left) An example of optimal continuations found for a given execution of a 2D reaching trajectory. The reproduction must smoothly approach a given perturbed goal while avoiding the obstacle. (Middle) An example of optimal adaptive continuations found for a given execution of a 2D reaching trajectory. The reproduction smoothly changes with the changing endpoint. (Right) A comparison of ELTE and DMPs for an endpoint changing partially through execution.}
    \label{fig:elte_obs_sim}
\end{figure}

We first validate our trajectory generation approach in a simulated 2D reaching environment. The results are shown in Fig.~\ref{fig:elte_obs_sim} (left). In this experiment, a demonstration is given which approaches a given target while avoiding an obstacle. An obstacle avoidance constraint is included in the optimization problem. The target in this experiment moves around its initial endpoint, while still avoiding the obstacle. Several possible continuations of the trajectory are shown to various novel positions around the endpoint. Each continuation smoothly continues to approximate the shape of the trajectory before approaching the desired endpoint. Additionally, all continuations maintain smoothness across the current point, avoiding discontinuities which could cause jerk in the robot execution.

Additionally, we examine a continuously changing endpoint in simulation. As shown in Fig.~\ref{fig:elte_obs_sim} (middle), a demonstration is given, and a reproduction is found for the demonstrated endpoint. During execution, the endpoint moves to a new location, adapting the reproduction. Again, the endpoint is changed, and a new adaptation is found. The time at which the adaptation is made is shown using opacity, with higher opacities adapting later in the execution. This shows that our method can adapt online to a continuously changing endpoint. However, because of the continuously changing endpoint the adaptation struggles to maintain the shape of the demonstrated trajectory. For continuous perturbations, predictive adaptation of perturbations could be used to better reproduce a perturbed trajectory.

As shown in Fig.~\ref{fig:elte_obs_sim} (right), we also compare our trajectory generation against Dynamic Movement Primitives (DMPs)~\cite{pastorDMP2009}. This demonstration has sharp corners that DMPs do not meet exactly. However, ELTE reproduces corners correctly, maintaining the shape which is important for tasks such as writing or welding where corners must be met exactly to successfully complete the task.

\subsection{Experimental Setup}

\begin{figure}[ht]
    \centering
    \includegraphics[width=0.389\linewidth]{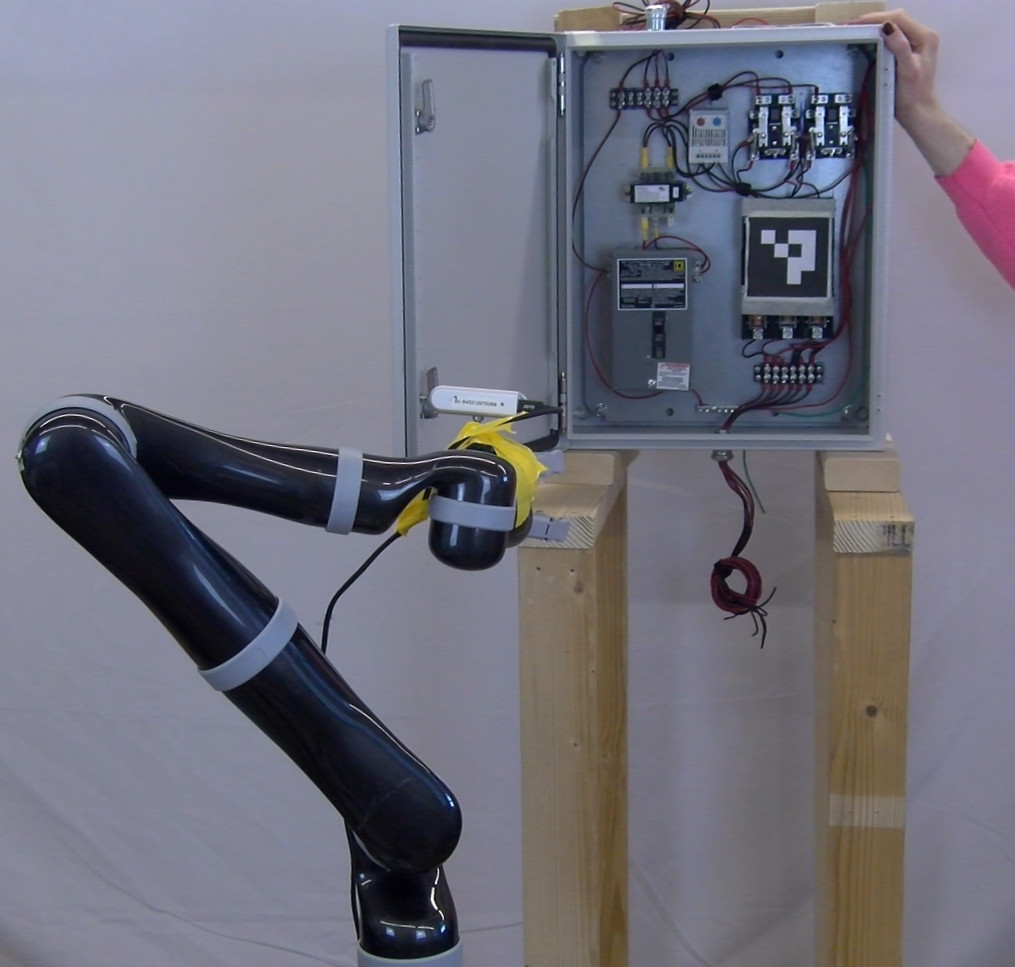}
    \includegraphics[width=0.22\linewidth]{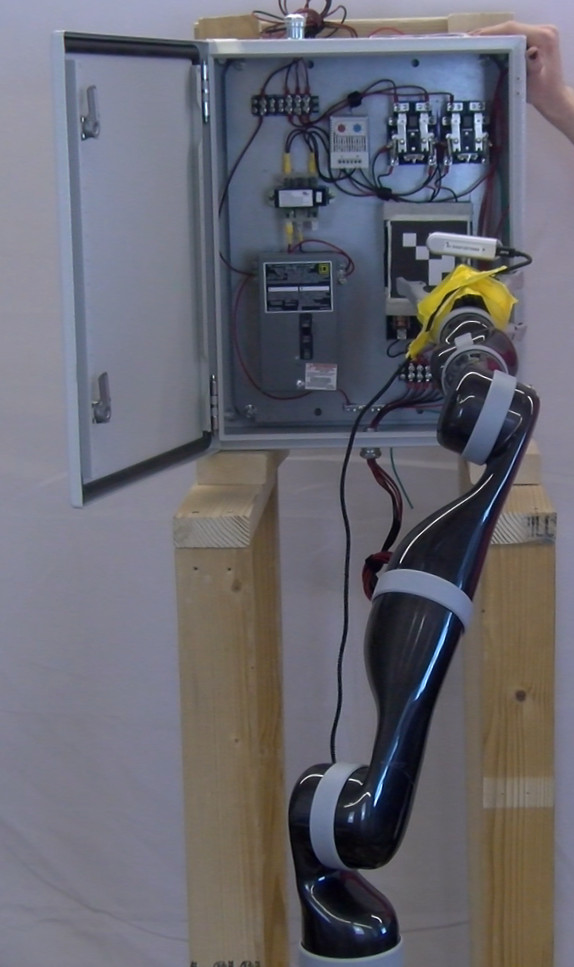}
    \includegraphics[width=0.36\linewidth]{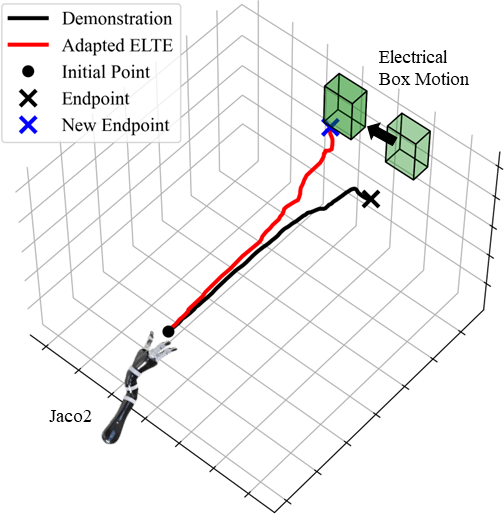}
    \caption{\small Execution of an inspection task before (left) and after (center) the environment changes and the target is moved. Our framework reacts appropriately to the change in the environment and successfully executes the task (right).}
    \label{fig:box_adapt}
\end{figure}

We validated our approach in several real world experiments. Our robotic platform consists of the Kinova Jaco2 7DOF manipulator arm and the Ghost Robotics Vision 60 (V60) legged robot. We mounted the Jaco2 on the V60 using a custom-designed and fabricated plate (shown in Fig~\ref{fig:setup}). In the first set of experiments, the goal was to approach an AR marker within an electrical box (shown in Fig~\ref{fig:box_adapt}). A camera was affixed to the end-effector for inspection of the electrical box. A demonstration using kinesthetic teaching was taken with this setup without perturbations in the base or target. The electrical box was placed on a cart with wheels such that it may be moved around during task execution. The V60 remained standing during these experiments, but did move slightly due to the shifting payload.

Additionally, we used this system on a dynamic moving platform shown in Fig.~\ref{fig:setup}. Our moving platform consists of a wooden base atop a Motek M-Gait treadmill with the ability for generating periodic pitch and sway motions. Connected to this platform, we mounted an AR marker to represent the electrical box for inspection. The camera remained affixed to the robot end-effector. We experimented with the V60 standing and stepping in place. During the stepping in place trials, a human user teleoperated the V60 to keep it in the center of the platform. Note that we do not consider communication between the Jaco2 and V60 in any experiment. The V60 is running the default off-the-shelf walking controller made for stable joystick control and does not implement obstacle avoidance or gait control. The Jaco2 runs our proposed framework. For all experiments, the parameters used for the sigma-points  of the UKF were set to $\alpha = 1$, $\beta = \num{2e-6}$, and $\kappa = 0$. The parameters for the adaptive skill learning were $w_E = w_R = 0.001$. Skill reproductions were generated in task space and executed using closed-form inverse kinematics with a low-level controller. Orientations during execution are generated using slerp~\cite{Shoemake1985slerp}.

\subsection{Real-world Experiments}
\label{subsec:rlexp}

We first validate our approach in an environment with a dynamically moving target. In these experiments, shown in Fig.~\ref{fig:box_adapt}, the electrical box is moved by a human to simulate a dynamic surface. The base is not controlled and remains mostly still, but there is slight movement as it balances in response to the shifting weight of the arm. Fig.~\ref{fig:box_adapt} (right) shows the original demonstration given as well as the reproduced trajectory. The demonstration given was with a still base on a higher platform, therefore the arm has to adapt initially to reaching up, then adapts to the movement of the electrical box over the course of execution. The box is moved to the left during execution, and the reproduction adapts accordingly as shown in the adapted trajectory in Fig.~\ref{fig:box_adapt}. This validates our framework in a real-world dynamic environment, where the arm is still able to approach the target successfully.

We perform experiments where our framework is expected to react to real-world movements in real-time using the moving platform setup described in the previous section. We test using a dynamic moving platform with pitch and sway motions at various speeds, and with the V60 either standing or stepping in place for various levels of instability. We compare motions with and without adaptation. The light motion is programmed to have the treadmill pitch $\pm$3\degree at 1.5 Hz and with a sway of $\pm$0.5 m at 2 Hz. The heavy motion is programmed to have the treadmill pitch $\pm$8\degree and sway $\pm$0.5 m both at 2.4 Hz. The results of this are shown in Table~\ref{tab:treadmill-table} and the accompanying video\footnote{Accompanying video: \url{https://youtu.be/H342Y0Hxl_0}}. In this table, \textit{S} for \textit{Success} denotes tests in which the robot successfully reaches the box, \textit{F} for \textit{Fail} denotes tests which do not successfully reach the box, for instance because of poor adaptation or an emergency stop intervention to prevent unsafe execution, and \textit{P} for \textit{Partial Fail} denotes tests which still complete the task but had issues during execution, such as obstacle collisions (obstacle avoidance constraints are not included in this experiment). For light or no motion, there is no significant difference in performance with and without the adaptive framework. However, in heavier motions, especially when the base is more unstable due to stepping in place, the performance with our framework is more successful. The proactive adaptation is able to track the moving target and predict its motion, and the reactive adaptation stops unsafe execution, preventing hard fails.

\begin{table}
\centering
\caption{\small{Results of testing with and without the proactive-reactive framework (PRF) for the dynamic surface with varying levels of motion (\textit{S}: Success, \textit{P}: Partial fail, \textit{F}: Fail)}}
\addtolength{\tabcolsep}{-0.52em}
\begin{tabular}{ll|ccc|ccc} \toprule
& & \multicolumn{3}{c|}{Without PRF} & \multicolumn{3}{c}{With PRF}
\\\cmidrule(lr){3-5}\cmidrule(lr){6-8}
           & & Test 1  & Test 2 & Test 3    & Test 1  & Test 2 & Test 3\\\midrule
\multirow{2}{*}{No Motion} & Standing & \textit{S} & \textit{S} & \textit{S} & \textit{S} & \textit{S} & \textit{S} \\
 & Step in place & \textit{S} & \textit{S} & \textit{S} & \textit{S} & \textit{S} & \textit{S} \\
\multirow{2}{*}{Light Motion} & Standing & \textit{S} & \textit{S} & \textit{S} & \textit{S} & \textit{S} & \textit{S} \\
  & Step in place & \textit{S} & \textit{S} & \textit{P} & \textit{S} & \textit{S} & \textit{S} \\
\multirow{2}{*}{Heavy Motion} & Standing  & \textit{S} & \textit{S} & \textit{S} & \textit{S} & \textit{S} & \textit{S} \\
 & Step in place & \textit{P} & \textit{F} & \textit{F} & \textit{S} & \textit{P} & \textit{P} \\ \bottomrule
\end{tabular}

\label{tab:treadmill-table}
\end{table}

\begin{figure}[ht]
    \centering
    \includegraphics[width=0.98\linewidth, height=5cm]{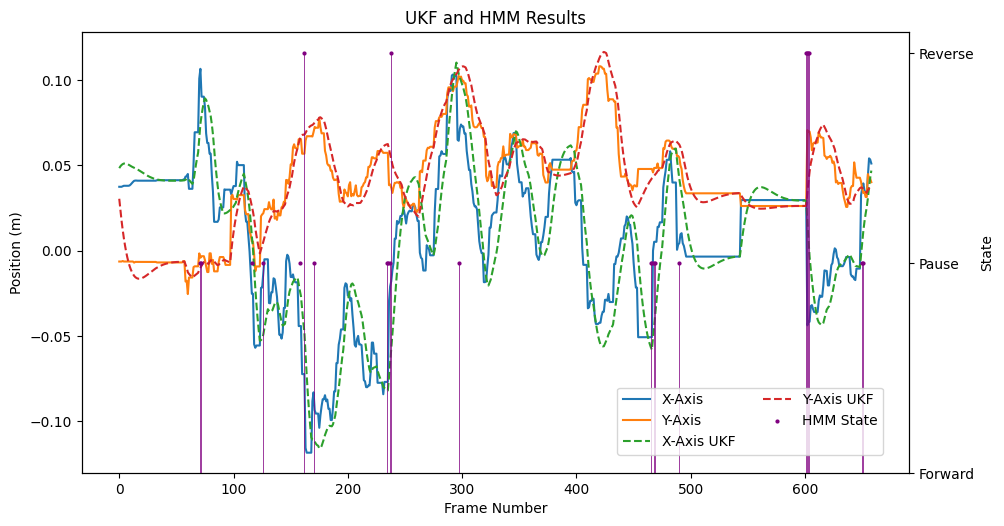}
    \caption{\small X and Y position of the target relative to the camera, the UKF predicted output of these positions, and the HMM output state during execution. Results are for a run with light motion on the treadmill while the V60 is stepping in place.}
    \label{fig:reactive_plots}
\end{figure}

The performance of the reactive system is shown in Fig.~\ref{fig:reactive_plots}, where the UKF and HMM results are shown for a run in the configuration where the base is stepping in place, and the platform is moving with light motion. The UKF is able to track the position of the tag to provide a reliable short term prediction of the motion of the AR tag. Additionally, the HMM provides state transitions when the tag begins to become unstable, and the robot reacts appropriately by pausing or reversing execution, and continues only when stability returns.

Finally, we compare our LfD method, Elastic-Laplacian Trajectory Editing (ELTE), with Dynamic Movement Primitives (DMPs)~\cite{pastorDMP2009}. The framework shown in Fig.~\ref{fig:framework} is used for both methods, only the skill learning module is changed. Both execute on the moving platform while the legged base is stepping in place. As reported in Table~\ref{tab:deviations-table}, ELTE provides a more stable execution as it consistently has a lower average distance from target and a lower standard deviation of the distance. This indicates better tracking and adaptation ability, as the target was kept in the camera frame with more stability. 

\begin{table}
\centering
\caption{\small{The mean and standard deviation of the distance (mm) between the target and the center of the camera when comparing ELTE and DMP methods in varying levels of surface motion.}}
\addtolength{\tabcolsep}{-0.4em}
\begin{tabular}{lcccc} \toprule
& \multicolumn{2}{c}{DMP} & \multicolumn{2}{c}{ELTE}
\\\cmidrule(lr){2-3}\cmidrule(lr){4-5}
           & mean distance & std & mean distance & std\\\midrule

No Motion & 124.84 & 42.51 & 112.99 & 37.34 \\            
Light Motion & 117.61 & 48.07 & 107.44 & 34.03 \\         
Heavy Motion & 130.67 & 56.94 & 125.75 & 51.24 \\         \bottomrule
\end{tabular}

\label{tab:deviations-table}
\end{table}


\section{Conclusions and Future Work}
\label{sec:conclusion}

In this work we propose and validate an adaptive skill learning framework for manipulation in dynamic environments, as well as design and test a novel Learning from Demonstration (LfD) representation for adaptive skill learning and generation in unstructured environments. This framework combines skill adaptation with a reactive system for safe execution, while the LfD representation provides fast and smooth adaptation which maintains the shape of a given demonstration. We evaluate the validity of our framework through 36 real-world experiments using a legged mobile manipulator under a variety of disturbances. 
Our framework is shown to provide safer execution compared to trials without any reactive behaviors, and our LfD representation provides more robust and stable adaptation compared to other adaptive LfD representations.

There are a myriad of opportunities for future work in this under-explored field of manipulation in dynamic environments. Firstly, creating more proactive adaptation is a possible avenue. Here, we use an Unscented Kalman Filter for state prediction, but other filters could be used or possibly a neural network trained for predictive movements. Additionally, in the case where the manipulator is attached to a dynamic base, investigating full-body control for proactive or reactive movements may yield better results.

\section*{Acknowledgements}
This work was supported in part b  the U.S. Office of Naval Research (N00014-21-1-2582 and N00014-23-1-2744) and National Science Foundation (FRR-2237463).
\typeout{}
\bibliographystyle{IEEEtran}
\bibliography{references}

\end{document}